%% file: acl_latex.tex
\pdfoutput=1

\documentclass[11pt]{article}

\usepackage[preprint]{acl}

\usepackage{times}
\usepackage{latexsym}

\usepackage[T1]{fontenc}

\usepackage[utf8]{inputenc}

\usepackage{microtype}

\usepackage{inconsolata}

\usepackage{graphicx}

\usepackage{todonotes}

\title{Inductive Bias Extraction and Matching for LLM Prompts}

\author{ %
\textbf{Christian M. Angel, Francis Ferraro} \\
  Department of Computer Science and Electrical Engineering\\
  University of Maryland, Baltimore County\\
  Baltimore, MD 21250 USA \\
  \texttt{\{change1,ferraro\}@umbc.edu} \\
  }

\begin{document}
\maketitle
\begin{abstract}
    The active research topic of prompt engineering makes it evident that LLMs are sensitive to small changes in prompt wording. A portion of this can be ascribed to the inductive bias that is present in the LLM. By using an LLM's output as a portion of its prompt, we can more easily create satisfactory wording for prompts. This has the effect of creating a prompt that matches the inductive bias in model. Empirically, we show that using this Inductive Bias Extraction and Matching strategy improves LLM Likert ratings used for classification by up to 19\% and LLM Likert ratings used for ranking by up to 27\%.
\end{abstract}

\section{Introduction}


If you wanted to build a patio, you may ask an LLM about how to get started. Two of the steps that it could suggest are "Purchase patio blocks," and "Place patio blocks in the desired area." While placing patio blocks is a logical course of action after purchasing patio blocks, it does not necessarily follow that you must place them immediately, or you may want to build something else that also requires patio blocks. This kind of out of the ordinary application may confuse an LLM that has only seen these two segments in this order, and would be liable to label the placement of the patio blocks as a result of purchasing.

To determine the relationship between two text segments (such as in the patio block example), it will be helpful to know whether one is a direct result of the other, or if the one is a requirement of the other. These requirement and result relationships are subtly different, and it would be helpful to have a fine grained scale for both to help quantify what the relationship is. The issue is that producing high quality scales of this nature would require a knowledge of the inductive biases of the LLM for that particular task. Normally this would be determined through a lengthy prompt engineering process, but creating a method for automatically extracting these inductive biases would require less labor and would not be subject to human error.

\begin{figure}
    \centering
    \includegraphics[width=0.9\linewidth]{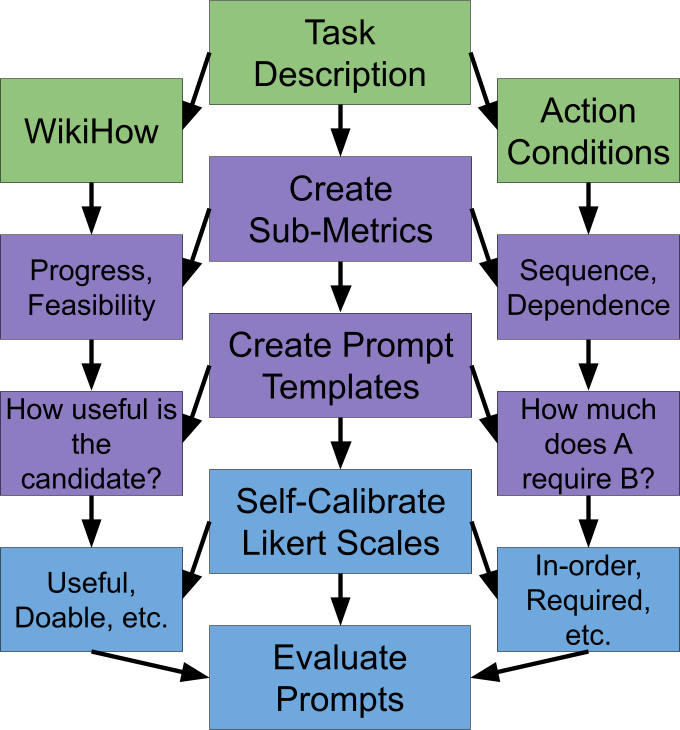}
    \caption{An overview of the IBEaM pipeline. Inputs are in green, human steps are in purple, and LLM steps are in blue.}
    \label{fig:overview}
\end{figure}


The method that we use to improve LLMs' performance on non-comparative classification tasks is a process that we call Inductive Bias Extraction and Matching (IBEaM), shown in Figure \ref{fig:overview}. By incorporating prior knowledge we have about the task, namely what sub-tasks it can be broken down into and how we should combine them once they are solved, we can more fully take advantage of the different strengths of the LLM. When using the LLM to solve these sub-tasks, ideally we would know what the optimal prompt would be to solve the sub-task. As shown by the large body of work done in prompt engineering \cite{Chen_2025}, LLMs are sensitive to small changes in prompt wording, and a portion of this can be ascribed to the inductive bias that is present in the LLM that gives it a "preference" for certain wordings. Usually these issues with prompt wording are solved through a repetitive prompt engineering process, but by prompting the LLM for its preference for prompt wording, we can eliminate the need for using prompt engineering to find the specific wording that works best with an LLM's inductive biases. IBEaM is not inherently limited to non-comparative classification tasks, but due to the difficulty of quantifying performance changes in other tasks, we limit the scope of this paper to this type of task.


For each of the tasks that we examine in this paper, we prompt an LLM for one or more 10 point Likert scales that can be used to generate one or more scores for each instance for each task. These scores are combined to generate an overall score (or multiple overall scores if the task is multiple-choice) that is used to perform classification. For each task, we also evaluate a simple baseline where the LLM is prompted for a rating from 1 to 10 (or, again, multiple ratings if the task is multiple-choice) for each instance. Like the scores generated with our IBEaM method, these ratings are used to perform the downstream classification task.


Our contributions are:
\begin{itemize}
    \item We define methods for inductive bias extraction and including that extracted inductive bias in prompts to improve LLM performance.
    \item We demonstrate that you can describe the criteria to evaluate and let the LLM identify its preferred wording.
    \item We show that an LLM's ability to generate numeric scores is limited, and we demonstrate that IBEaM can be used to generate improved numeric scores from LLMs.
    \item We perform ablation studies that show that all components of IBEaM are useful.


\end{itemize}


\section{Related Work}


With the advent of generative large language models (LLMs) \cite{achiam2023gpt, touvron2023llamaopenefficientfoundation, brown2020language}, much research has been conducted to determine what tasks they can be applied to and what their efficacy is for those tasks \cite{zhao2025surveylargelanguagemodels}. While LLMs exhibit emergent behaviors that allow them to perform tasks that machine learning methods were previously ineffective for, they can also be applied to tasks that existed prior to LLMs being introduced. The motivation for applying LLMs to these older tasks may be the desire to improve efficiency in execution speed, improve accessibility to non-technical users, or improve the state-of-the-art in accuracy. This broad spectrum of LLM applications is what has motivated us to evaluate IBEaM by generating numeric scores and using them as a proxy for performing classification and ranking tasks.

One method of improving LLM performance on tasks is to perform chain of thought reasoning, where the LLM is walked through the process of solving one or more examples in the prompt prior to being asked to evaluate an instance \cite{NEURIPS2022_9d560961}. Another method of improving LLM performance is to perform prompt engineering \cite{Chen_2025}, which is typically a repetitive process that optimizes an LLM's downstream performance by making small adjustments to its prompt. All prompt engineering techniques fundamentally are attempting to make the prompt work better with the parameters learned by the LLM, or in other words, are attempting to find the prompt that is most compatible with the LLM's inductive biases.

All of the datasets examined in this paper have some concept of a world state implied by the text. LLMs are able to reason about the hypothetical world state implied in a prompt, and also the real world state implied by their training data \cite{zhu2023large}. The encoding of this real world state is a form of inductive bias in LLMs.

The category of task that we examine in this paper is classification. \citet{gretz-etal-2023-zero} selected several classification tasks and LLMs and compared the LLMs' zero-shot performance to their performance when fine tuned. \citet{sun-etal-2023-text} also created a novel prompting technique to improve LLMs' ability to perform text classification.

Using LLMs to produce Likert ratings has been evaluated by \citet{chiang-lee-2023-closer}. In their work, they found that asking an LLM to explain the rating that it assigned improves performance over methods that constrain the LLM output to a single rating integer. We apply this technique of explanation generation to IBEaM to optimize our performance.

The concept of breaking a problem into more manageable sub-problems has also been researched previously. LLM-Eval is a system for improving LLM output by breaking a problem down into subtasks and prompting an LLM to fill a schema for the response \cite{lin-chen-2023-llm}. There are some interesting technical differences between LLM-Eval and our work. First, LLM-Eval uses schema filling, but we found that, in our particular prompt formulation for IBEaM, schema filling was less effective than freeform text generation. Second, LLM-Eval uses a single prompt to evaluate multiple metrics, but again for our particular prompt formulation for IBEaM, we found that using multiple prompts for the different metrics was most effective.

There is also previous work on using a multi-step conversational approach with LLMs. \citet{trivedi-etal-2023-interleaving} proposed a technique called Interleaving Retrieval with Chain-of-Thought (IRCoT). With their approach, they use a sequence of prompts to perform chain-of-thought reasoning, but between each of the prompts, they query a knowledge database. The information retrieved from the database is used to influence the next prompt in the sequence. While we perform something similar with our sequence of prompts in IBEaM, we do not query an external knowledge source. Instead, the intermediate information that we use is obtained from the LLM.

\section{Method}

\begin{figure}
    \centering
    \includegraphics[width=\linewidth]{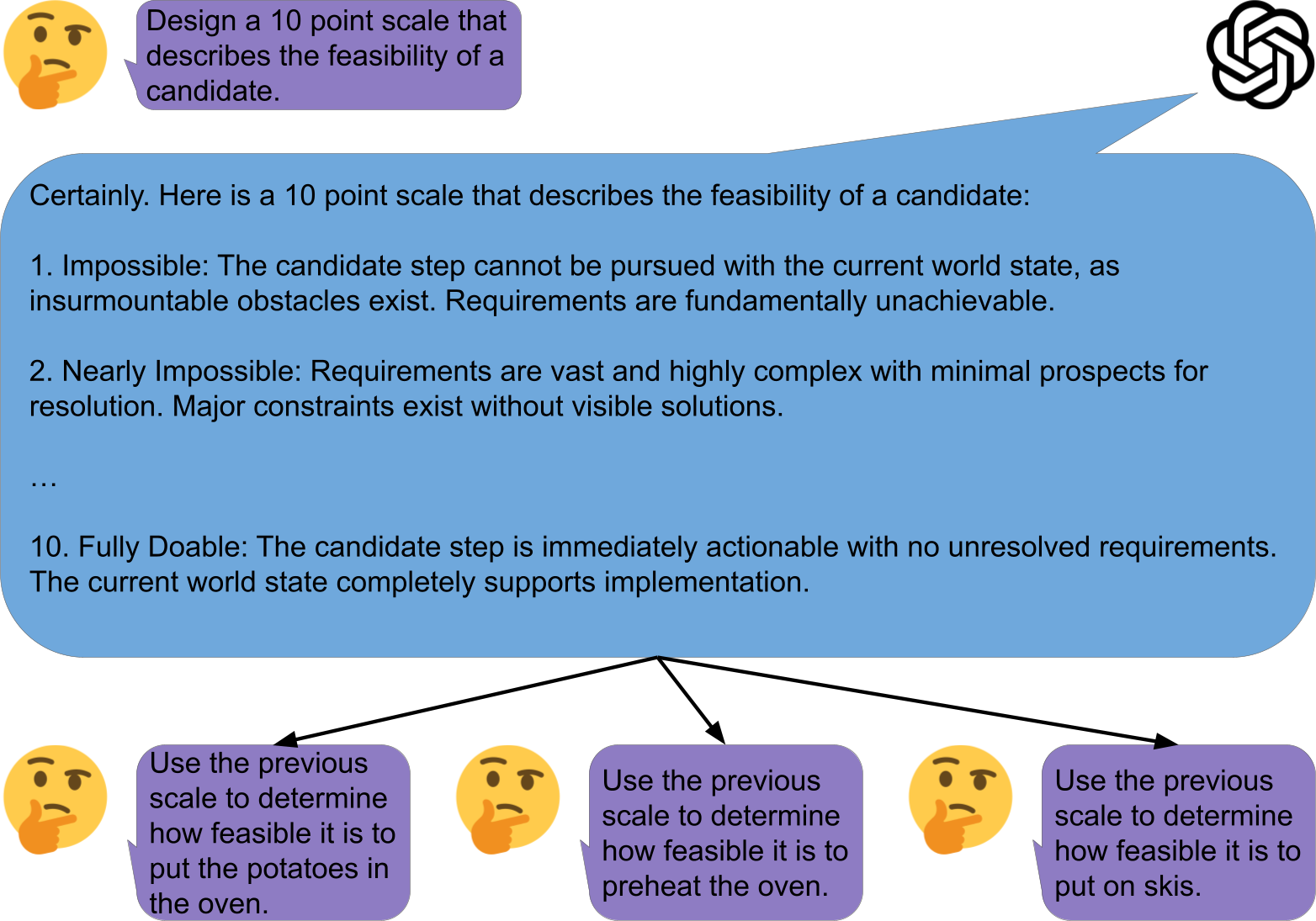}
    \caption{For each of our trials, we generate a Likert scale for each metric and apply it to all instances in that trial. 
    We include the scale as part of the conversation history when making future calls to the LLM. Keeping the scale the same for all instances improves rating consistency and reduces the cost and computation time when compared to regenerating scales for each instance.}
    \label{fig:conversation_history}
\end{figure}

\begin{figure*}
    \centering
    \includegraphics[width=\linewidth*9/10]{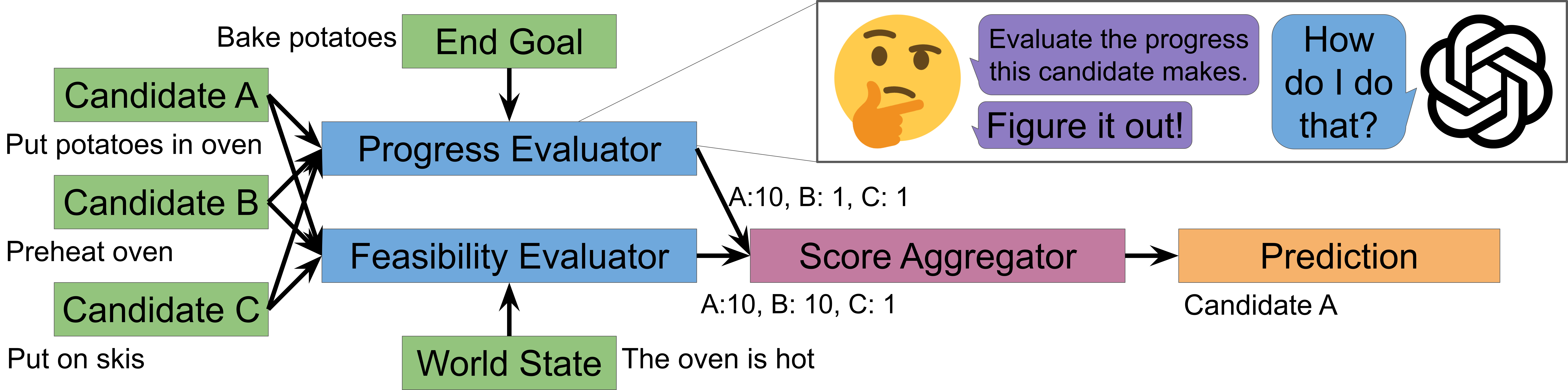}
    \caption{An overview of IBEaM in use for our WikiHow task. As in Figure \ref{fig:overview}, green squares indicate input from the dataset. The core of the pipeline is the set of evaluators designed for the given task. For WikiHow, we have a progress evaluator and a feasibility evaluator. The output from these evaluators is given to an appropriate aggregator for the task, which then makes the final prediction for the instance. Similar pipelines that were designed for SAGA Task 2 and the Action Conditions task are shown in Figures \ref{fig:ibeam-saga} and \ref{fig:ibeam-action-conditions}, respectively, in Appendix \ref{sec:ibeam_pipelines}.}
    \label{fig:wikihow}
\end{figure*}

Applying IBEaM to a new task is a four step process. First, the user must create some number of component metrics to enable the extraction of additional inductive bias. Second, the user must create a prompt template that can integrate the extracted inductive bias into an evaluation prompt. Third, the user must prompt the LLM to extract the inductive bias for each component metric, and finally, the user must devise some method of combining the component scores into a final evaluation. Once all of these steps have been performed, the result is a pipeline that can be used to improve the performance of an LLM on the desired task.

\subsection{Creation of Component Metrics}

The first step in the process of applying IBEaM to a problem is to break the problem into a reasonable number of fine-grained sub-problems. The goal here is not to break the problem into overly small problems, but rather to determine the high-level dimensions that define the task. For example, if our task is to determine how well "Putting a potato in the oven" continues the process of "Baking a potato" given a previously completed step of "Preheat the oven," two logical dimensions of this task are progress and feasibility. Ideally, when determining whether or not a step is a logical continuation of a process when working towards a goal, we want the candidate step to both make substantial progress and to be substantially feasible. Other tasks will break into different component metrics, and this is a determination that needs to be made by the IBEaM user.

Splitting a task into component sub-tasks is not unique to IBEaM, but IBEaM obtains outsized benefits from this process: the more Likert scales obtained from an LLM across unique dimensions, the more inductive bias can be extracted from the LLM. The benefit that IBEaM gets from this additional inductive bias extraction is in addition to the intuitive benefits from breaking a task into logical sub-problems. These intuitive benefits include getting more detailed output from the LLM, getting an evaluation across multiple dimensions, and, if applicable, getting the ability to unevenly weight the answers to the different sub-problems.

\subsection{Inductive Bias Extraction}

The next step in the process of applying IBEaM to a problem is to extract the LLM's inductive bias for each component metric. For all of the tasks that we study in this paper, this involves prompting the LLM for a Likert scale for each component metric. For example, for the feasibility metric mentioned previously, the LLM may produce a 10 point scale along the lines of the scale seen in Figure \ref{fig:conversation_history}. As the figure shows, we can extract the inductive bias of the model in the form of its preferred wording for each rating in the Likert scale. If we were to attempt to create this scale manually as opposed to prompting the LLM for it, we would have to perform a labor intensive prompt engineering process.

\subsection{Inductive Bias Matching}

The third step in the process of applying IBEaM to a problem is to match the LLM's inductive bias when prompting. This involves creating a prompt to evaluate each metric which includes the generated Likert scale. As we demonstrate with our experiments, the prompts for evaluating each metric do not need to be complicated, and can include the generated Likert scale as part of the LLM's conversation history, which allows us to skip any postprocessing of the Likert scales in our experiments. This conversation history is utilized in the prompt by using wording such as, "Rate the following candidate from 1 to 10 on the previous feasibility scale." A simplified version of this process is shown in Figure \ref{fig:conversation_history}.

An additional benefit of this inductive bias matching process is that we obtain a consistent rating scale that we can apply to all instances in a task, which allows the LLM to apply consistent and more specific criteria when compared to intuitive baseline approaches.

\subsection{Combine Component Metrics}

As a result of splitting a task into multiple sub-problems, we have the issue of there being multiple component scores that need to be combined in some way to make an overall score. The exact method by which we combine the scores varies depending on the type of task. For ranking tasks like WikiHow, a simple summation of the scores is usually an effective and intuitive starting point. We tested summation, product, summation of score ranks, and product of score ranks. Of these we found that the product of score ranks was the most effective combination method while not requiring any learned parameters. Assuming $n$ is the number of multiple choice options, this method works by assigning a rank from 0 to $n-1$ to each candidate for each metric. The final combined score is the product of these ranking numbers. The benefit of this method is that it promotes candidates that have more balanced scores compared to the other candidates while still promoting candidates that have high scores overall.

Our other two tasks, SAGA Task 2 and the Action Conditions task, are classification tasks as opposed to ranking tasks. We decided to use logistic regression as a score aggregator for these tasks, and the justification for this is in Appendix \ref{logistic_regression_justification}.

\section{Datasets}

We use three different datasets to evaluate the efficacy of IBEaM when compared to intuitive baselines. These datasets include SAGA, the Action Conditions dataset, and our own WikiHow-based dataset. SAGA and the Action Conditions dataset allow us to evaluate IBEaM's ability to enable classification while our WikiHow-based dataset allows us to evaluate IBEaM's ability to perform ranking.

\subsection{WikiHow Procedural Dataset}


We wanted to evaluate IBEaM's effect on the LLM's reasoning capabilities, so WikiHow was selected as one of our datasets because performing tasks with it requires an understanding of implied world state. We created a 5-way multiple choice task from WikiHow where each instance contains an in-progress goal, some number of completed steps, and five candidate continuations for the procedure. The five candidates are composed of the target step and four distractors. The four distractors include a duplicate step that has already been performed, two out-of-document steps, and an out-of-order future step. The out-of-order future step is part of the same article as the goal and the target, but there is another step that needs to be completed first before the out-of-order future step can be performed. Of these distractors, the duplicate step is difficult to rank correctly because it is usually feasible to repeat a step (e.g. preheat the oven) when it already has been done, but it does not make any progress towards the goal. However, the out-of-order future step is even more difficult because it makes progress towards the goal (e.g. leave the potatoes in the oven for an hour), but the LLM needs to identify the gap between its requirements (e.g. potatoes in the oven) and the world state (e.g. potatoes on the table) in order to determine that it is not a viable continuation to the current process.

We annotated 66 instances with the target, distractors, previously completed steps, and overall goal. Since a training set is not required for our ranking method, this dataset is exclusively a test set and does not contain training or validation splits.

\subsection{Summary Applicability Dataset}

The ROCStories dataset is composed of instances referred to as stories, which have five events in a sequence that can be summarized by a single sentence \cite{mostafazadeh-etal-2016-corpus}. The PASTA dataset is an extension of ROCStories that substitutes alternative events into the stories, which may or may not match up with the original summary sentence \cite{ghosh2023pastadatasetmodelingparticipant}. The Story Alternatives and Goal Applicability (SAGA) dataset is an annotated extension to PASTA, which among other things, contains human evaluations of how applicable these alternative stories in PASTA are to the original summary \cite{vallurupalli2024sagaparticipantspecificexaminationstory}. The paper for SAGA defines multiple tasks that can be tested on SAGA, and in particular we will be performing Task 2, which is a binary classification task that tests a model's ability to determine if a summary is applicable to a story or not. We chose this dataset because it is nontrivial to perform, but it also has the simple evaluation criteria of Macro and Micro F1 score.

\subsection{Pre/Post-Condition Classification Dataset}

The Action Conditions dataset is also derived from WikiHow \cite{wu-etal-2023-learning}, but instead of focusing on ranking next steps in a process, it instead defines segments of text within WikiHow articles as pre-conditions and post-conditions. In this context, a pre-condition relationship indicates that one text segment is a requirement for the other, while a post-condition relationship indicates that one text segment is the result of another. Any pair of text segments within a WikiHow article that does not have an indicated pre-condition or post-condition relationship has what they refer to as a NULL relationship. These relationships are not necessarily bidirectional, so if one text segment is a pre-condition of another, the second text segment is not necessarily a pre-condition or post-condition of the first unless labeled as such. For our evaluations, we define an instance as a pair of text segments, a label, and the minimum body of text that contains both text segments. We chose this dataset because it is also nontrivial to perform, has simple evaluation criteria of Macro and Micro F1 scores, and is sufficiently different from our other two tasks.

\section{Experimental Setup}

For all of our IBEaM experiments, each trial uses a freshly generated set of Likert scales. Both IBEaM and the baselines re-prompt the LLM even in cases where the prompt would be identical to a prompt from another trial. The reason for this is that we observed that the LLM tended to have small variances in their responses even when given an identical prompt. Within each trial, LLM responses are cached to reduce inference time and cost. The LLMs used for both IBEaM and the baselines is the 2024-08-06 snapshot of GPT-4o \cite{hurst2024gpt} and LLaMa 3.2 \cite{grattafiori2024llama3herdmodels}.

\subsection{WikiHow Step Candidate Ranking}

For IBEaM, we split the task into progress and feasibility sub-tasks. Each of these sub-tasks has a Likert scale generated for it which is then included as part of the conversation history for the LLM. Each Likert scale is regenerated for each trial, but is consistent for each instance within a trial. The most effective score aggregator that we found was to rank the scores for progress and feasibility individually, and then take the product of the two numeric rankings. Whichever candidate has the highest product is selected as the prediction.


For our baseline, we prompt the LLM for a rating from 1 to 10 for each candidate given the in-progress goal and the completed steps. The predicted candidate is the one with the highest rating.

\subsection{Summary Classification Task}

For IBEaM, we split the task into summarization, objective, and accomplishment sub-tasks. Like the WikiHow task, each of these subtasks has a Likert scale regenerated for it for each trial. However, unlike the WikiHow task, SAGA Task 2 is a binary classification task as opposed to a ranking task, so we cannot rank the scores to determine what the best prediction is. Instead, we use a separate set of instances no larger than the test set to fit a logistic regression model. This logistic regression model takes in the numeric ratings as one-hot encoded vectors and predicts either true or false for applicability, and we fit separate logistic regression models for IBEaM and the baseline. The complete pipeline is shown in Figure \ref{fig:ibeam-saga} in Appendix \ref{sec:ibeam_pipelines}.

The test set for SAGA is composed of 512 instances, and each trial uses the same test set, of these 512 instances, 419 have a positive label while only 93 have a negative label. This imbalance in labeling is representative of the training set, so we apply a modifier to the class weights when fitting the logistic regression model. A different 512 instance sample is taken from the training set for each trial to fit the logistic regression model.

For our baseline, we again prompt the LLM for a rating from 1 to 10 for how applicable a goal is to a sequence of steps. Like the method we used for IBEaM, we again fit a logistic regression model to determine which ratings predict true or false for applicability. For each trial, the training set used to fit the logistic regression model for the baseline is the same one that is used for IBEaM.

\subsection{Pre/Post-Condition Classification Task}

For IBEaM, we split the task into sequence, dependence, and independence sub-tasks. Like the WikiHow task and SAGA Task 2, each of these sub-tasks has a Likert scale regenerated for it for each trial. The Action Conditions task is a ternary classification task as opposed to a binary classification task, but we can still apply the same logistic regression method that we used for SAGA Task 2. Again, we get a number of training instances for our logistic regression model, but instead of predicting true or false, it predicts pre-condition, post-condition, or NULL. The complete pipeline is shown in Figure \ref{fig:ibeam-action-conditions} in Appendix \ref{sec:ibeam_pipelines}.

The test set for the Action Conditions dataset is far too large to evaluate in its entirety, so instead we take a different 512 instance sample from the test set for each trial. Like SAGA Task 2, a different 512 instance sample is taken from the training set for the purposes of fitting the logistic regression models for each trial.

For our baseline, we prompt the LLM for two different scores on a scale from 1 to 10. The first score is a pre-condition score and the second score is a post-condition score. Like SAGA Task 2, these two scores are given to a trained logistic regression model to determine whether the prediction should be pre-condition, post-condition, or NULL. For each trial, the training set and the test set for the baseline are the same ones that are used for IBEaM.

\section{Results}

\begin{table}[t]
\resizebox{\linewidth}{!}{%
\begin{tabular}{cl|cc|cc}
      & \multicolumn{1}{c|}{} & \multicolumn{2}{c|}{Accuracy} & \multicolumn{2}{c}{MRR}         \\
      & \multicolumn{1}{c|}{} & IBEaM            & Baseline   & IBEaM          & Baseline       \\ \hline
      & Trial 1               & \textbf{57.3\%}   & 50.0\%      & \textbf{0.718} & 0.598          \\
      & Trial 2               & \textbf{59.1\%}   & 49.5\%      & \textbf{0.751} & 0.610          \\
GPT   & Trial 3               & \textbf{63.6\%}   & 49.2\%      & \textbf{0.771} & 0.605          \\
4o    & Trial 4               & \textbf{58.3\%}   & 47.7\%      & \textbf{0.728} & 0.561          \\
      & Trial 5               & \textbf{61.4\%}   & 44.2\%      & \textbf{0.764} & 0.560          \\
      & Mean                  & \textbf{59.9\%}   & 48.1\%      & \textbf{0.746} & 0.587          \\ \hline
      & Trial 1               & \textbf{26.8\%}   & 26.0\%      & 0.472          & \textbf{0.476} \\
      & Trial 2               & \textbf{32.3\%}   & 19.8\%      & \textbf{0.513} & 0.423          \\
LLaMa & Trial 3               & \textbf{23.8\%}   & 17.3\%      & \textbf{0.464} & 0.380          \\
3.2   & Trial 4               & \textbf{27.3\%}   & 17.6\%      & \textbf{0.498} & 0.405          \\
      & Trial 5               & \textbf{27.9\%}   & 21.5\%      & \textbf{0.505} & 0.417          \\
      & Mean                  & \textbf{27.6\%}   & 20.4\%      & \textbf{0.490} & 0.420         
\end{tabular}%
}
\caption{An overview of the performance of IBEaM and the baseline for each WikiHow trial. Accuracy values give partial credit for ties. If the target has the same score as one distractor, half credit is given, two distractors result in one third credit, and so on.}
\label{tab:wikihow_results}
\end{table}

\begin{figure}[t]
    \centering
    \includegraphics[width=\linewidth]{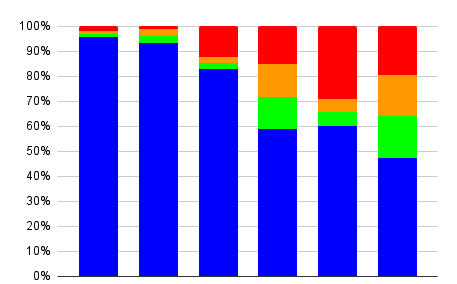}
    \includegraphics[width=\linewidth]{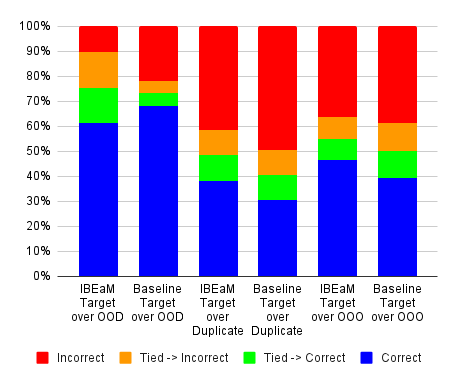}
    \caption{A breakdown of IBEaM and the baseline's performance on the WikiHow dataset. Performance when using GPT-4o and LLaMa 3.2 are shown in the top and bottom figures, respectively. Assigned scores are integers, so ties are split evenly between correct and incorrect predictions.} 
    \label{fig:wikihow_breakdown}
\end{figure}

For all of our experiments, we see at least a small improvement in the relevant evaluation metrics on average. The amount of improvement varies widely depending on the task, and that can likely be attributed to the LLMs' ability to perform well with the baseline prompt. In particular, the LLMs struggles to perform well on the Action Conditions baseline, which leads to the largest increase in performance when applying IBEaM.

\subsection{WikiHow Step Candidate Ranking}

As shown in Table \ref{tab:wikihow_results}, we show an almost 12 percentage point improvement in accuracy and more than a 0.15 improvement in MRR over the baseline when averaged across all of the GPT-4o trials. LLaMa showed a 7 percentage point improvement in accuracy and a 0.7 improvement in MRR. For this dataset, all trials show consistent improvement over the baseline as well, with the exception of the MRR score for one of the LLaMa 3.2 trials. We also performed an analysis of the performance on each individual distractor to see if each category of distractor was as difficult as we expected. As shown in Figure \ref{fig:wikihow_breakdown}, the performance on each category is about what we expected for both IBEaM and the baseline. Of particular interest is the number of scoring ties produced by each method. The GPT-4o baseline, likely as a result of its predictions being derived from a single 10 point scale, had many ties when evaluating both the duplicate distractor and the out-of-order distractor. IBEaM had substantially fewer ties when applied to GPT-4o, and this was likely in part because it was combining multiple metrics, which allows for finer grained scoring predictions than the baseline. Even with its increased certainty, it still outperformed the baseline in all categories both when counting all ties as incorrect predictions and when counting 50\% of ties as incorrect predictions. Curiously, these patterns do not hold when IBEaM is applied to LLaMa 3.2, and instead has many more ties than the baseline for OOD instances and a comparable number of ties for duplicates and OOO instances.

\subsection{Summary and Pre/Post-Condition Classification Tasks}

As shown in Table \ref{tab:classification_results}, out of the five SAGA Task 2 trials we performed, all F1 scores showed improvement for GPT-4o and LLaMa 3.2, with the exception of one micro F1 score when IBEaM was applied to GPT-4o. For Trial 4 where the Micro F1 score was lower than the baseline, we need to recall that there is a substantial labeling imbalance in the test set, meaning that if IBEaM or the baseline is more biased towards labeling instances as applicable, the Micro F1 score will increase and the Macro F1 score will decrease. The inverse relationship is true as well, so this is an indication that IBEaM and the baseline performed about the same overall in Trial 4. However, the other four trials show a consistent increase in both Macro and Micro F1 scores when compared to the baseline, so IBEaM performs better overall.


\begin{table}[t]
\centering
\resizebox{\linewidth}{!}{
\begin{tabular}{ccc|cc}
           &       &         &                      & IBEaM                  \\
           &       &         &                      & Improvement            \\
           &       &         & Baseline             & over Baseline          \\ \hline
           &       & Trial 1 & 67.4 / 61.6          & \textbf{4.30 / 4.44}   \\
           &       & Trial 2 & 64.3 / 59.5          & \textbf{5.47 / 4.42}   \\
           & GPT   & Trial 3 & 64.3 / 58.5          & \textbf{1.95 / 2.27}   \\
           & 4o    & Trial 4 & \textbf{68.9 / 62.0} & -1.56 / 1.40           \\
           &       & Trial 5 & 65.6 / 60.3          & \textbf{8.01 / 4.11}   \\
SAGA       &       & Mean    & 66.1 / 60.4          & \textbf{3.63 / 3.33}   \\ \cline{2-5} 
Task 2     &       & Trial 1 & 48.0 / 45.9          & \textbf{18.36 / 10.64} \\
           &       & Trial 2 & 43.2 / 42.6          & \textbf{12.11 / 7.91}  \\
           & LLaMa & Trial 3 & 51.4 / 48.8          & \textbf{9.57 / 4.95}   \\
           & 3.2   & Trial 4 & 43.8 / 42.9          & \textbf{16.41 / 9.78}  \\
           &       & Trial 5 & 47.7 / 45.7          & \textbf{10.16 / 5.65}  \\
           &       & Mean    & 46.8 / 45.2          & \textbf{13.32 / 7.78}  \\ \hline
           &       & Trial 1 & 46.5 / 40.5          & \textbf{0.39 / 1.77}   \\
           &       & Trial 2 & 44.7 / 42.0          & \textbf{2.93 / 1.13}   \\
           & GPT   & Trial 3 & 40.2 / 36.2          & \textbf{13.09 / 10.45} \\
           & 4o    & Trial 4 & 45.5 / 38.7          & \textbf{4.10 / 5.42}   \\
           &       & Trial 5 & 32.8 / 29.9          & \textbf{18.55 / 18.01} \\
Action     &       & Mean    & 42.0 / 37.5          & \textbf{7.81 / 7.36}   \\ \cline{2-5} 
Conditions &       & Trial 1 & 34.0 / 32.5          & \textbf{2.15 / 2.63}   \\
           &       & Trial 2 & 30.3 / 28.8          & \textbf{12.89 / 10.66} \\
           & LLaMa & Trial 3 & 28.5 / 27.8          & \textbf{9.18 / 7.48}   \\
           & 3.2   & Trial 4 & \textbf{42.4 / 35.5} & -5.86 / -2.53          \\
           &       & Trial 5 & 34.6 / 32.0          & \textbf{8.59 / 6.05}   \\
           &       & Mean    & 33.9 / 31.3          & \textbf{5.39 / 4.86}  
\end{tabular}
}
\caption{An overview of the performance of IBEaM and the baseline for the classification tasks. F1 scores are multiplied by 100 and expressed as Micro F1 / Macro F1. Improvement is calculated with (IBEaM F1 - Baseline F1). When using GPT-4o, IBEaM consistently performs better than the baseline with the exception of one SAGA Task 2 trial. When using LLaMa 3.2, IBEaM consistently performs better than the baseline with the exception of one Action Conditions trial.}
\label{tab:classification_results}
\end{table}

For the Action Conditions task, IBEaM performs better than the baseline in all trials for GPT-4o and four out of five trials for LLaMa 3.2, but considering that the Action Conditions task is a ternary classification problem, both IBEaM and the baseline struggle to perform well on this task.

\section{Ablation Studies}

To determine how much each component of IBEaM contributes to the overall performance improvements, we perform two ablation studies. In the first ablation study, we remove the self-calibration of the Likert scales from IBEaM, but we still have the split metrics. This helps us determine the degree to which the metric splitting improves the performance, which should be nonzero because it allows us to extract more inductive bias from the model. In the second ablation study, we remove the split metrics from IBEaM, but we still perform self-calibration. This helps us determine how much the self-calibration improves the performance, which also should be nonzero because that is the key step where the inductive bias matching occurs.

\subsection{Removing Self-Calibration}

As shown in Table \ref{tab:saga_task2_ablation_results}, we see that removing the Likert scale calibration step from IBEaM has a strong negative effect on the average accuracy in that it reduces the average accuracy to less than that of the baseline. However, a notable effect of removing the calibration step is that it also reduces the standard deviation of the accuracies across the trials to be below that of both unablated IBEaM and the baseline. While this is a positive effect, it does not justify the reduction in accuracy when compared to the baseline.

\begin{table}[t]
\centering
\resizebox{\linewidth}{!}{
\begin{tabular}{c|cccc}
        & IBEaM                & Uncal.               & Unsplit              & Baseline    \\ \hline
Trial 1 & \textbf{71.7 / 66.0} & 67.0 / 60.8          & 70.1 / 64.9          & 67.4 / 61.6 \\
Trial 2 & \textbf{69.7 / 63.9} & 66.6 / 59.6          & 60.9 / 58.2          & 64.3 / 59.5 \\
Trial 3 & 66.2 / 60.7          & 67.2 / 61.1          & \textbf{71.5 / 66.8} & 64.3 / 58.5 \\
Trial 4 & 67.4 / 63.4          & 63.7 / 58.4          & \textbf{72.9 / 66.5} & 68.9 / 62.0 \\
Trial 5 & \textbf{73.6 / 64.4} & 63.3 / 59.5          & 59.2 / 56.7          & 65.6 / 60.3 \\
Mean    & \textbf{69.7 / 63.7} & 65.5 / 59.9          & 66.9 / 62.6          & 66.1 / 60.4 \\
SD      & 3.04 / 1.92          & \textbf{1.91 / 1.10} & 6.36 / 4.81          & 2.05 / 1.45
\end{tabular}
}
\caption{An overview of the ablation studies we performed for SAGA Task 2 with GPT-4o. F1 scores are multiplied by 100 and expressed as Micro F1 / Macro F1. IBEaM without split metrics performs well in the majority of trials, but has the highest standard deviation. Uncalibrated IBEaM (with split metrics) has the lowest SD, but it does not outperform the baseline in the majority of trials. Unablated IBEaM has a higher SD than uncalibrated IBEaM, but its average increase in accuracy makes it a worthwhile tradeoff.}
\label{tab:saga_task2_ablation_results}
\end{table}

\subsection{Removing Metric Splitting}

Looking again at Table \ref{tab:saga_task2_ablation_results}, we see that removing the metric splitting does reduce the average accuracy when evaluating SAGA Task 2. However, what is more notable is that the standard deviation of the accuracy across the trials more than doubles when compared to unablated IBEaM. While IBEaM without split metrics is more accurate than the baseline on average, having triple the standard deviation of the baseline is undesirable.

\subsection{Analysis}

Our two ablation studies show that both the metric splitting and the self-calibration within IBEaM are necessary for it to achieve consistently high accuracies on our tasks. It is notable that the self-calibration step, on average, increases the accuracy of IBEaM to be more than that of the baseline, but causes a substantial increase in standard deviation. On the other hand, the metric splitting has a strong moderating effect on the variance of the accuracies across the trials both with and without self-calibration. When both components are combined, unablated IBEaM is more accurate than both ablated versions while having a reasonable standard deviation, in spite of the uncalibrated IBEaM having a lower average accuracy than the baseline. This shows us that the components within IBEaM complement each other, and this synergy makes IBEaM more than the sum of its parts.

\section{Conclusion}

In this paper, we presented IBEaM, which is a method for writing prompts that automatically extract and apply the LLM's inductive biases in a way that improves its ability to assign scores that can be used for downstream ranking and classification. This ability of LLMs to extract their own inductive biases likely has broad applications, but we quantified the possible performance improvements in multiple contexts with concrete evaluation criteria. When compared to our baselines that prompt the LLM for a rating from 1 to 10 with a short description of the task, the increased complexity of IBEaM is a worthwhile investment in contexts where a numeric score is required in addition to an overall prediction of ranking or classification.

\section*{Limitations}

All of our experimentation were performed using GPT-4o, which is a closed source LLM trained on a proprietary dataset, and LLaMa 3.2, which is an open source LLM. IBEaM is not dependent specifically on GPT-4o or LLaMa 3.2, but it is a technique specifically designed for interacting with LLMs, and as such, is subject to the same limitations as whatever LLM it is being used with.



\bibliography{acl_latex.bib}

\appendix

\section{Logistic Regression as a Score Aggregator}
\label{logistic_regression_justification}

Using IBEaM on a classification task requires us to create some sort of decision boundary depending on the values of the scores. Empirically, we found that feeding one-hot encoded vectors for all of the sub-metrics into logistic regression was the most effective way of creating a decision boundary. This, however, necessitates the use of a training set for the logistic regression model which is not required for ranking tasks. As such, for SAGA Task 2 and the Action Conditions task, we take a small sample of the training set and feed that through the IBEaM pipeline. The labels and the LLM's responses are then used to fit the logistic regression models for both of these tasks.

While the logistic regression component is not a part of IBEaM, it does aid in evaluating the improvement in the ratings produced by IBEaM over the ratings produced by the baseline. As a result of us using the LLM to produce ratings rather than direct classification predictions, we need a way to convert the ratings into a binary classification. Logistic regression does this for us while still being informative because it will perform better when the ratings are closer to being linearly separable. Whichever prompting strategy that produces more linearly separable results is the prompting strategy that is more certain about its predictions. While certainty is not a direct proxy for downstream results, using a separate training set for logistic regression will also translate any decreased quality in the ratings to decreased accuracy.

\section{IBEaM Pipelines}
\label{sec:ibeam_pipelines}

\begin{figure*}
    \centering
    \includegraphics[width=\linewidth]{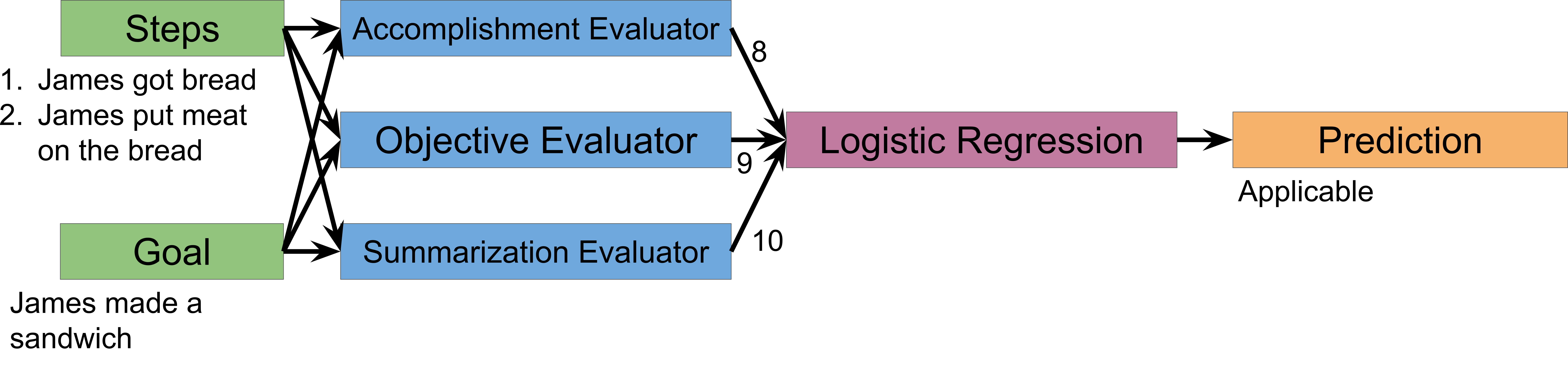}
    \caption{An overview of IBEaM in use for SAGA Task 2.}
    \label{fig:ibeam-saga}
\end{figure*}

\begin{figure*}
    \centering
    \includegraphics[width=\linewidth]{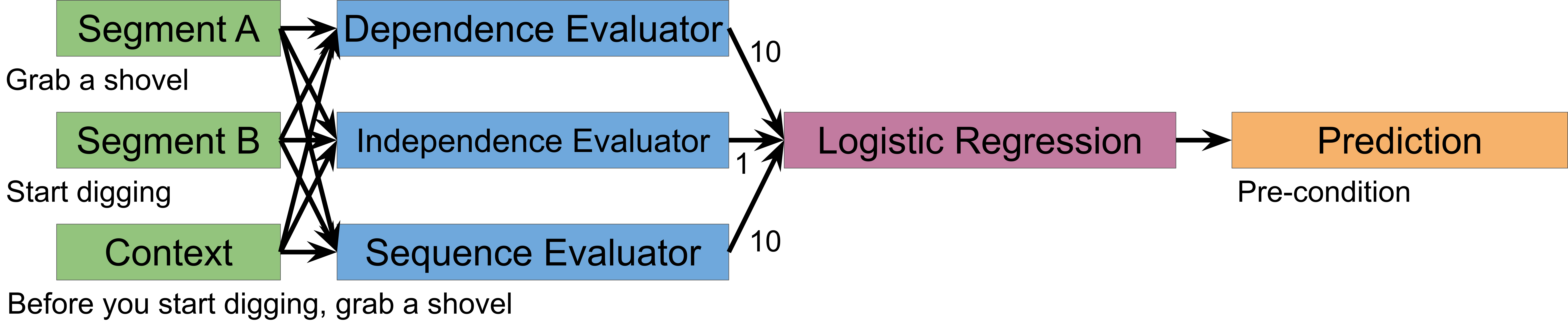}
    \caption{An overview of IBEaM in use for the Action Conditions Task.}
    \label{fig:ibeam-action-conditions}
\end{figure*}

Figures \ref{fig:ibeam-saga} and \ref{fig:ibeam-action-conditions} show the pipeline setup we used to evaluate SAGA Task 2 and the Action Conditions task, respectively.

\section{Full Classification Results}

Table \ref{tab:classification_results_expanded} shows an expanded version of Table \ref{tab:classification_results} with a direct comparison of the baseline and IBEaM's F1 scores.

\begin{table*}[]
\centering
\resizebox{\textwidth}{!}{
\begin{tabular}{cl|cccc|cccc}
      &         & \multicolumn{4}{c|}{SAGA Task 2}                                                     & \multicolumn{4}{c}{Action Conditions}                                                  \\
      &         & \multicolumn{2}{c|}{Micro F1}                        & \multicolumn{2}{c|}{Macro F1} & \multicolumn{2}{c|}{Micro F1}                        & \multicolumn{2}{c}{Macro F1}    \\
      &         & IBEaM          & \multicolumn{1}{c|}{Baseline}       & IBEaM            & Baseline   & IBEaM          & \multicolumn{1}{c|}{Baseline}       & IBEaM          & Baseline       \\ \hline
      & Trial 1 & \textbf{0.717} & \multicolumn{1}{c|}{0.674}          & \textbf{0.660}   & 0.616      & \textbf{0.469} & \multicolumn{1}{c|}{0.465}          & \textbf{0.423} & 0.405          \\
      & Trial 2 & \textbf{0.697} & \multicolumn{1}{c|}{0.643}          & \textbf{0.639}   & 0.595      & \textbf{0.477} & \multicolumn{1}{c|}{0.447}          & \textbf{0.431} & 0.420          \\
GPT   & Trial 3 & \textbf{0.662} & \multicolumn{1}{c|}{0.643}          & \textbf{0.607}   & 0.585      & \textbf{0.533} & \multicolumn{1}{c|}{0.402}          & \textbf{0.467} & 0.362          \\
4o    & Trial 4 & 0.674          & \multicolumn{1}{c|}{\textbf{0.689}} & \textbf{0.634}   & 0.620      & \textbf{0.496} & \multicolumn{1}{c|}{0.455}          & \textbf{0.441} & 0.387          \\
      & Trial 5 & \textbf{0.736} & \multicolumn{1}{c|}{0.656}          & \textbf{0.644}   & 0.603      & \textbf{0.514} & \multicolumn{1}{c|}{0.328}          & \textbf{0.479} & 0.299          \\
      & Mean    & \textbf{0.697} & \multicolumn{1}{c|}{0.661}          & \textbf{0.637}   & 0.604      & \textbf{0.498} & \multicolumn{1}{c|}{0.420}          & \textbf{0.448} & 0.375          \\ \hline
      & Trial 1 & \textbf{0.664} & \multicolumn{1}{c|}{0.480}          & \textbf{0.565}   & 0.459      & \textbf{0.361} & \multicolumn{1}{c|}{0.340}          & \textbf{0.351} & 0.325          \\
      & Trial 2 & \textbf{0.553} & \multicolumn{1}{c|}{0.432}          & \textbf{0.505}   & 0.426      & \textbf{0.432} & \multicolumn{1}{c|}{0.303}          & \textbf{0.394} & 0.288          \\
LLaMa & Trial 3 & \textbf{0.609} & \multicolumn{1}{c|}{0.514}          & \textbf{0.537}   & 0.488      & \textbf{0.377} & \multicolumn{1}{c|}{0.285}          & \textbf{0.353} & 0.278          \\
3.2   & Trial 4 & \textbf{0.602} & \multicolumn{1}{c|}{0.438}          & \textbf{0.526}   & 0.429      & 0.365          & \multicolumn{1}{c|}{\textbf{0.424}} & 0.329          & \textbf{0.355} \\
      & Trial 5 & \textbf{0.578} & \multicolumn{1}{c|}{0.477}          & \textbf{0.514}   & 0.457      & \textbf{0.432} & \multicolumn{1}{c|}{0.346}          & \textbf{0.380} & 0.320          \\
      & Mean    & \textbf{0.601} & \multicolumn{1}{c|}{0.468}          & \textbf{0.530}   & 0.452      & \textbf{0.393} & \multicolumn{1}{c|}{0.339}          & \textbf{0.362} & 0.313         
\end{tabular}
}
\caption{An overview of the performance of IBEaM and the baseline for the classification tasks. When using GPT-4o, IBEaM consistently performs better than the baseline with the exception of one SAGA Task 2 trial. When using LLaMa 3.2, IBEaM consistently performs better than the baseline with the exception of one Action Conditions trial.}
\label{tab:classification_results_expanded}
\end{table*}

\section{LLM Conversations}
\label{sec:llm_conversations}

Figures \ref{fig:llm_conversation_feasibility}, \ref{fig:llm_conversation_progress}, and \ref{fig:llm_conversation_baseline} show sample conversations for the WikiHow task.

\begin{figure*}

\input{llm_conversation_feasibility.tex}

\caption{A conversation with GPT-4o to determine how feasible a step is for the WikiHow task.}
\label{fig:llm_conversation_feasibility}

\end{figure*}

\begin{figure*}

\input{llm_conversation_progress.tex}

\caption{A conversation with GPT-4o to determine the progress that a step makes for the WikiHow task.}
\label{fig:llm_conversation_progress}

\end{figure*}

\begin{figure*}

\input{llm_conversation_baseline.tex}

\caption{An example baseline prompt and response for the WikiHow task.}
\label{fig:llm_conversation_baseline}

\end{figure*}

\end{document}

%% file: llm_conversation_feasibility.tex
\subparagraph{User\\}

Imagine you are in a world where you are working towards some goal and
you have already done some number of steps. Design a 10 point scale that
describes how doable some candidate step is. On this scale, 10 should
represent the highest degree of doability and 1 should represent the
lowest degree of doability.

\bigskip

\subparagraph{Assistant\\}

Creating a 10-point scale to assess the doability of a candidate step in
achieving a goal requires careful consideration of various factors such
as feasibility, resources, time, skills, and potential barriers. Here's
a proposed scale:

\begin{enumerate}
\def\labelenumi{\arabic{enumi}.}
\item
  \textbf{1 - Not Doable:} The step is virtually impossible with current
  resources and knowledge. Major constraints exist with insurmountable
  obstacles, requiring significant change in circumstances or assets.
\item
  \textbf{2 - Highly Unlikely:} The step faces substantial barriers and
  requires resources or skills that are currently inaccessible. It would
  need a drastic shift or breakthroughs to proceed.
\item
  \textbf{3 - Very Difficult:} The step is achievable only under
  specific and rare conditions. Requires significant allocation of time
  and effort, and is likely dependent on external factors beyond
  control.
\item
  \textbf{4 - Challenging:} Requires overcoming several hurdles; some
  might be ambiguous or unknown. The step demands significant planning,
  effort, and possible learning of new skills or resource acquisition.
\item
  \textbf{5 - Moderate Difficulty:} The step is doable with effort but
  involves dealing with known obstacles. Requires clear planning and
  moderate resources. Some risk or uncertainty is involved.
\item
  \textbf{6 - Fairly Doable:} The step is achievable with diligent work
  and problem-solving. Challenges and resource needs are identified,
  with some manageable risks.
\item
  \textbf{7 - Doable:} The step is straightforward with existing
  resources. Expected challenges are routine but can be mitigated with
  careful planning and execution.
\item
  \textbf{8 - Highly Doable:} Readily achievable given current abilities
  and resources. Minor obstacles may need attention, but they're
  manageable within the normal scope of work.
\item
  \textbf{9 - Very Highly Doable:} The step is easily actionable with
  existing tools and skills. Minimal challenges or resistance expected.
\item
  \textbf{10 - Effortless:} The step can be completed smoothly and
  quickly with available resources and competencies. No foreseeable
  challenges; outcome is nearly guaranteed.
\end{enumerate}

This scale takes into account factors such as resource availability,
skill level, time constraints, external dependencies, and potential
challenges, providing a comprehensive assessment of a step's feasibility
in the pursuit of a goal.

\end{figure*}
\begin{figure*}

\subparagraph{User\\}

Imagine you are in a world where your objective is to Assemble your
paper bag so you can make a paper bag and you have done the following:
\begin{itemize}
    \item Place the cut out paper in front of you on a flat surface.
    \item Fold the bottom edge of the paper up 2 inches (5 cm) and sharply crease the fold.
    \item Locate the center points of the top and bottom edges.
\end{itemize}

For the following query, do not assume that the requirements of the
following step are fulfilled unless they are known to be so from the
world state, and start your response with a rating from 1 to 10 on the
previous doability scale that considers the gap between the current
world state and the requirements of the following step: Fold the sides
of the bag into place.

\bigskip

\subparagraph{Assistant\\}

\textbf{Rating: 5 - Moderate Difficulty}

Assessing the doability of the next step, ``Fold the sides of the bag
into place,'' involves several considerations based on the current world
state:

\begin{enumerate}
\def\labelenumi{\arabic{enumi}.}
\item
  \textbf{Understanding of Task:} Assuming you know where the sides
  should be folded based on the shape and design of your cut-out paper
  bag, your familiarity with this task affects doability.
\item
  \textbf{Requirements and Preparation:} The paper is on a flat surface,
  and you've folded the bottom edge and located the center points.
  However, it's not specified whether additional markings or guidelines
  for side folds are present.
\item
  \textbf{Resources and Tools:} The task seems to involve manual
  folding, which doesn't require additional resources beyond what's
  mentioned. If the paper is pre-scored or has guidelines, it simplifies
  the task.
\item
  \textbf{Skill Level:} Folding paper requires basic skills, but
  achieving precise and clean folds can be challenging without
  experience or guidelines, affecting the neatness and structural
  integrity of the bag.
\item
  \textbf{Potential Barriers:} Without specific fold lines or additional
  instructions, there might be some trial and error involved, which
  could complicate the task.
\end{enumerate}

Based on these considerations, folding the sides involves moderate
difficulty due to potential lack of precise guidelines and the manual
dexterity required for accurate folds, though it's achievable with
effort and focus.

%% file: llm_conversation_progress.tex
\subparagraph{User\\}

Imagine you are in a world where you are working towards some goal and
you have already done some number of steps. Design a 10 point scale that
describes the amount of progress that some candidate step makes. On this
scale, 10 should represent the highest degree of progress and 1 should
represent the lowest degree of progress.

\bigskip

\subparagraph{Assistant\\}

Creating a 10-point scale to assess the progress of a candidate step
towards a goal involves defining specific criteria for each level.
Here's a potential scale:

\begin{enumerate}
\def\labelenumi{\arabic{enumi}.}
\item
  \textbf{Minimal Progress (1/10)}: The step contributes negligible
  advancement. It's more of a repetition of previous actions with little
  to no incremental benefit.
\item
  \textbf{Very Low Progress (2/10)}: The step introduces a small benefit
  or change, but its impact on the overall goal is minor and barely
  noticeable.
\item
  \textbf{Low Progress (3/10)}: The step leads to some progress, but it
  mostly addresses minor or peripheral aspects of the goal without
  significantly moving it forward.
\item
  \textbf{Below Average Progress (4/10)}: The step results in measurable
  progress, yet it doesn't substantially alter the current status or
  momentum towards the goal.
\item
  \textbf{Moderate Progress (5/10)}: The step provides a clear and
  positive contribution, representing a necessary but not groundbreaking
  advancement.
\item
  \textbf{Above Average Progress (6/10)}: The step offers solid progress
  by effectively addressing key aspects or overcoming notable obstacles
  related to the goal.
\item
  \textbf{Significant Progress (7/10)}: The step leads to a meaningful
  leap forward, resolving critical issues or laying groundwork for
  further advances.
\item
  \textbf{High Progress (8/10)}: The step results in significant gains,
  driving substantial momentum and bringing the goal within closer
  reach.
\item
  \textbf{Very High Progress (9/10)}: The step achieves major
  advancements and overcomes significant barriers, substantially
  transforming the outlook towards achieving the goal.
\item
  \textbf{Maximum Progress (10/10)}: The step represents an optimal
  move, producing transformative impact, and driving substantial
  completion or realization of the goal. It might even redefine the
  goal's parameters due to its effectiveness.
\end{enumerate}

This scale provides a structured way to assess and compare the efficacy
of different steps towards a specific goal, helping prioritize actions
and allocate resources efficiently.

\end{figure*}
\begin{figure*}

\subparagraph{User\\}

Imagine you are in a world where your objective is to Assemble your
paper bag so you can make a paper bag and you have done the following:
\begin{itemize}
    \item Place the cut out paper in front of you on a flat surface.
    \item Fold the bottom edge of the paper up 2 inches (5 cm) and sharply crease the fold.
    \item Locate the center points of the top and bottom edges.
\end{itemize}

For the following query, if the results of the following step have
already been done in the current world state, do not count that as
progress, and start your response with a rating from 1 to 10 on the
previous progress scale that considers the amount of progress that the
following step makes: Fold the sides of the bag into place.

\bigskip

\subparagraph{Assistant\\}

6/10

\textbf{Above Average Progress}: By folding the sides of the bag into
place, you are addressing a crucial step in the assembly process. This
action contributes effectively to the overall structure and stability of
the paper bag. It represents notable advancement from the preparatory
steps taken earlier, moving you closer to completing the functional
shape of the bag. This progress is solid and necessary, though other
steps will still be needed to achieve a fully assembled paper bag.

%% file: llm_conversation_baseline.tex
\subparagraph{User\\}

Imagine you are in a world where your objective is to Assemble your
paper bag so you can make a paper bag and you have done the following:
\begin{itemize}
    \item Place the cut out paper in front of you on a flat surface.
    \item Fold the bottom edge of the paper up 2 inches (5 cm) and sharply crease the fold.
    \item Locate the center points of the top and bottom edges.
\end{itemize}

For the following query, start your response with a rating from 1 to 10.
If you wanted to Assemble your paper bag so you can make a paper bag,
how good of a continuation is the following step? Fold the sides of the
bag into place.

\bigskip

\subparagraph{Assistant\\}

\begin{enumerate}
\def\labelenumi{\arabic{enumi}.}
\setcounter{enumi}{7}
\item
  Folding the sides of the bag into place is a logical next step after
  you've folded and creased the bottom edge and located the center
  points. This step is essential for forming the structure of the bag,
  but it could be more specific. For a higher rating, instructions could
  include using the center points to ensure symmetry and mentioning how
  to crease or fix the sides in place.
\end{enumerate}